\documentclass[10pt,twocolumn,letterpaper]{article}

\usepackage{svg}
\usepackage{multirow}
\usepackage[accsupp]{axessibility}
\usepackage{amsmath}
\usepackage{amssymb}
\usepackage{bbm}
\usepackage{xparse}
\usepackage{booktabs}
\usepackage{mathabx}
\usepackage{svg}
\usepackage{caption}
\usepackage{subcaption}

\newcommand{\mbp}{\mathbb{P}}
\newcommand{\mcp}{\mathcal{P}}
\newcommand{\mbq}{\mathbb{Q}}
\newcommand{\mcq}{\mathcal{Q}}

\NewDocumentCommand{\prob}{o m}{
    \mathbb{P}
    \IfValueT{#1}{_{#1}}
    \left( #2 \right)
}
\newcommand{\expec}[2]{\mathbb{E}_{#1} \left[ #2 \right]}

\usepackage[algorithms]{wacv}      

\usepackage{amsmath,amsfonts,bm}
\usepackage{upgreek}

\def\eqref#1{equation~\ref{#1}}

\def\1{\bm{1}}

\DeclareMathAlphabet{\mathsfit}{\encodingdefault}{\sfdefault}{m}{sl}
\SetMathAlphabet{\mathsfit}{bold}{\encodingdefault}{\sfdefault}{bx}{n}

\definecolor{wacvblue}{rgb}{0.21,0.49,0.74}
\usepackage[pagebackref,breaklinks,colorlinks,allcolors=wacvblue]{hyperref}

\def\wacvPaperID{26} 
\def\confName{WACV}
\def\confYear{2026}

\title{Detecting Object Tracking Failure via
Sequential Hypothesis Testing}

\author{Alejandro Monroy Muñoz, Rajeev Verma$^{*}$, Alexander Timans\footnote{*Equal advising}   \\
UvA-Bosch Delta Lab, University of Amsterdam\\
{\tt\small \{a.monroymunoz, r.verma, a.r.timans\}@uva.nl} \\
}

\begin{document}
\maketitle
\begin{abstract}
   Real-time online object tracking in videos constitutes a core task in computer vision, with wide-ranging applications including video surveillance, motion capture, and robotics. Deployed tracking systems usually lack formal safety assurances to convey when tracking is reliable and when it may fail, at best relying on heuristic measures of model confidence to raise alerts. To obtain such assurances we propose interpreting object tracking as a sequential hypothesis test, wherein evidence for or against tracking failures is gradually accumulated over time. Leveraging recent advancements in the field, our sequential test (formalized as an e-process) quickly identifies when tracking failures set in whilst provably containing false alerts at a desired rate, and thus limiting potentially costly re-calibration or intervention steps. The approach is computationally light-weight, requires no extra training or fine-tuning, and is in principle model-agnostic. We propose both supervised and unsupervised variants by leveraging either ground-truth or solely internal tracking information, and demonstrate its effectiveness for two established tracking models across four video benchmarks. As such, sequential testing can offer a statistically grounded and efficient mechanism to incorporate safety assurances into real-time tracking systems.
\end{abstract}    
\section{Introduction}
\label{sec:intro}
\begingroup
\let\thefootnote\relax
\footnotetext{$^*$Equal advising.}
\endgroup
Object tracking is a core competency in deployed vision systems, with applications in video surveillance \cite{video-surveillance}, motion analysis \cite{motion-analysis}, autonomous driving \cite{autonomous-driving, vehicle-navigation} or general traffic monitoring \cite{traffic-monitoring}. Unlike offline tracking which relies on access to complete video sequences, \emph{online} object trackers operate in real-time, and must ensure low-latency target localization given only past video frames. As such, object trackers inevitably encounter challenging settings and regularly fail due to occlusions, appearance changes, motion blur, or background clutter \cite{smith2005evaluating, luo2021multiple, ramachandra2020survey}. Despite substantial progress on improving the tracking models themselves \cite{object-tracking-survey}, reliably identifying \emph{if} and \emph{when} failures occur is left largely unaddressed. Existing approaches rely on heuristic estimates of model confidence, such as entropy or peak values of the tracker's internal response map \cite{tracking-failure-KCF, object-tracking-apce}, but lack formal statistical grounding. Therefore, system alerts may be missed, delayed, or raised unnecessarily often, prompting potentially costly false alerts
(\eg, by deferring to a human or triggering a more expensive re-calibration procedure).

\begin{figure}[t]
    \centering
    \includegraphics[width=\linewidth]{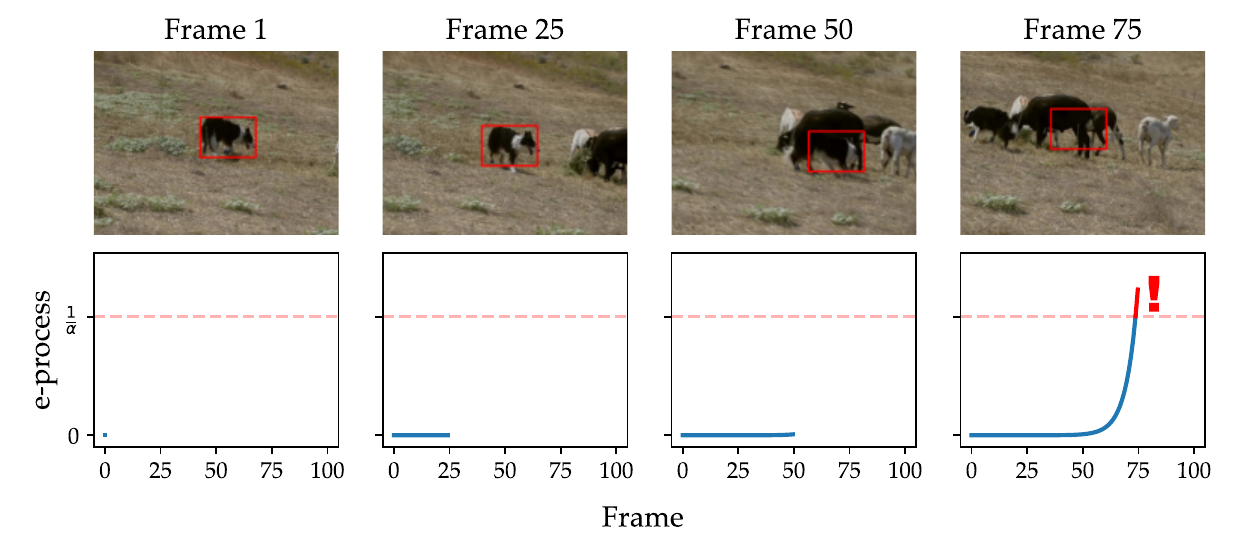}
    \caption{
    Conceptual example of a sequential test to detect object tracking failure. The tracker initially correctly follows the target object, but subsequently re-focuses on a wrong object due to occlusions. The test (e-process) mirrors the behaviour by remaining initially stable, but quickly grows and exceeds the signal threshold ($\frac{1}{\alpha}$) once tracking failure is determined, triggering an alert.
    }
    \label{fig:conceptual-example}
\end{figure}

To address this, we propose leveraging the framework of \emph{sequential hypothesis testing} \cite{e-book, merging-seq-e-values}, which offers statistical tools for \emph{anytime-valid} inference in sequential settings. Our particular design employs \emph{e-processes}, the sequential counterparts to more generic \emph{e-values} found in the wider literature on safe testing \cite{ramdas2023game, grunwald2024safe}, and which are tailored to data stream scenarios. E-processes permit repeated hypothesis testing with valid Type-I error control, regardless of when a decision is made (termed `optional stopping'). This renders them highly practical for online object tracking, where tracking failures may occur at any time step and real-time monitoring is desired. By framing the object tracking problem in terms of a verifiable hypothesis pair (\autoref{eq:hypothesis}), inherited Type-I error control properties provably contain a desired rate of false positives, equating a limit on unnecessarily triggered alerts. At the same time, our test design is ensured to accumulate per-frame evidence rapidly should a true failure occur, limiting the delay between occurrence and alert (see \autoref{fig:conceptual-example} for an illustration). We propose both supervised and unsupervised variants of the mechanism, depending on the availability of incoming information, which renders our approach highly flexible. Experiments with two established tracking models (via kernelized correlation filters and Siamese networks) across four video benchmarks (OTB-100, LaSOT, TrackingNet, and GOT-10k) assert that sequential testing offers a statistically grounded and efficient mechanism to embed formal assurances in object tracking systems. In summary, our contributions include:
\begin{itemize}
    \item Formulating object tracking failure as a sequential hypothesis testing problem, enabling principled failure detection with statistical guarantees (\autoref{subsec:method-problem-form});
    \item Designing an anytime-valid and efficient test that can operate in both supervised (via ground-truth bounding boxes) and unsupervised (via model-internal information) data stream scenarios (\autoref{subsec:method-tracking-metrics}, \autoref{subsec:method-bettingrate});
    \item Empirically validating the approach to assert contained false positive rates and low detection delays, and outlining important design choices and their effects (\autoref{sec:experiments}, \autoref{sec:results}).
\end{itemize}

\section{Related Work}
\label{sec:related}

\paragraph{Object tracking models.} \looseness=-1 Visual object tracking has evolved rapidly with the rise of deep learning, in particular seeing use of convolutional neural networks \cite{object-tracking-survey, luo2021multiple, danelljan2019atom}. Among model architectures, \emph{Siamese networks} such as SiamFC \cite{siamfc, chicco2021siamese} initiated a notable shift in tracking design. Search regions are analysed using a fully-convolutional architecture to match object templates, resulting in fast and generalizable tracking without the need for online adaptation. Recent advancements such as SiamRPN~\cite{li2018high} and STARK~\cite{yan2021learning} further integrate region proposal networks and transformer-based attention mechanisms, respectively. Relatedly, architectures such as TransTrack~\cite{sun2020transtrack, chen2021transformer} and OSTrack~\cite{ye2022joint} incorporate temporal long-range dependency modelling across video frames.

\looseness=-1 Kernelized Correlation Filters (KCF) \cite{kcf} and correlation-based tracking \cite{ma2015long, lukezic2017discriminative, bhat2018unveiling, zhen2020visual} enjoy high use for their generally efficient solutions, enabling real-time tracking on standard hardware. Although in part outperformed by more involved deep learning-based trackers in terms of accuracy, KCF-based tracking remains relevant in settings with strict computational constraints or desiring real-time responsiveness \cite{kumar2024correlation}.


\paragraph{Reliable failure detection.} \looseness=-1 Reliable detection of visual tracking failures is primarily addressed by means of heuristic scores. These are derived from model-internal confidences or response maps \cite{afat, object-tracking-apce, tracking-failure-KCF}, and compared to fixed thresholds to trigger alerts or re-calibration. The measures lack formal assurances, and can remain deceptively high (and thus unresponsive) even when the tracker has drifted, in particular for settings with visual clutter. Thus, recent work advocates for uncertainty-aware tracking, where uncertainty in localization or predictive distributions is explicitly modelled and propagated \cite{Yao_2025, lee2024uncertainty}, but remains heuristic. Earlier attempts at incorporating (non-sequential) hypothesis testing \cite{sheng2020hypothesis, enescu2007visual, kc2012iterative, demirbas2007maneuvering} consider involved graph- and modelling structures, and generally do not preoccupy themselves with safety assurances. 

\looseness=-1 Our proposal can be seen as complementary to any such heuristically-driven approaches, in that we may feed various model-internal information (\eg~correlation, uncertainty, or prediction signals) as an input to a sequential test, which `wraps' around the tracking pipeline to ensure statistical grounding via controlled false alerts (\autoref{eq:ville}). Our choice of framework aligns with other recent works employing sequential testing for more general model or data shift detection in stream settings \cite{e-detectors, vovk2021testing, timans2025continuous, podkopaev2021tracking, entropy-matching}, but which do not consider tracking applications.
\section{Detecting Tracking Failures via Testing}
\label{sec:method}

\looseness=-1 We next formalise our sequential testing framework by framing object tracking failure as a verifiable hypothesis pair, introducing our e-process design, and discussing both supervised and unsupervised tracking metrics.

\begin{figure*}[t]
	\includegraphics[width=\linewidth]{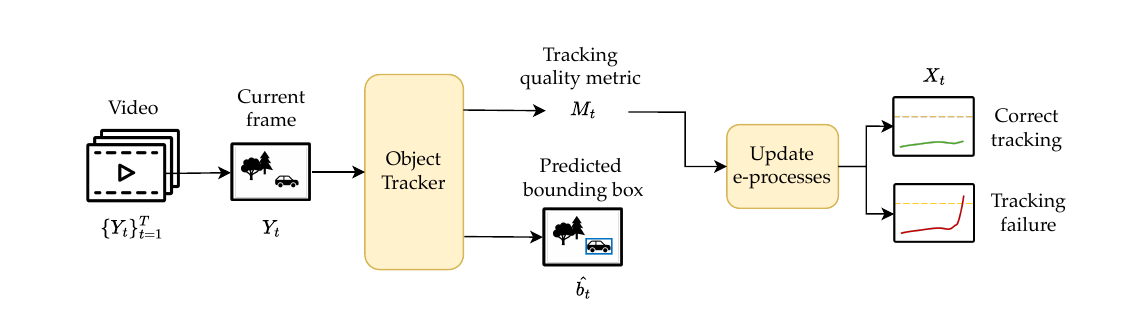}
	\caption{
    A schematic of the proposed sequential testing framework for object tracking. At any given time step $t$ the current frame $Y_t$ is fed to the tracker, whose produced prediction or response map (and potential ground truth information) inform a metric $M_t$ capturing object tracking quality. From it, evidence for or against tracking failure is added to a running measure of evidence (the e-process $X_t$, \autoref{eq:e-process}) whose growth either remains stable (when tracking is satisfactory) or eventually triggers a failure alert (when exceeding a threshold $\frac{1}{\alpha}$).
    }
	\label{fig:proposed-framework}
\end{figure*}

\subsection{Problem and Method Formulation}
\label{subsec:method-problem-form}

\looseness=-1 Let us denote an input video as a stream of frames $\{Y_t\}_{t=1}^T$, $Y_t \in \mathbb{R}^{W \times H \times C}$, with $W \times H$ the image dimensions and $C$ the number of channels (\eg, $C=3$ for RGB). A key requirement for our approach is that the user defines a bounded tracking metric $M_t \in [0,1]$ to assess tracking quality at every time step $t$, and a minimum fixed quality level $\epsilon \in (0, 1)$ \citep{aGRAPA}. Based on these parameters, our goal is to identify when tracking performance drops below the tolerance level $\epsilon$, which can be reformulated as the following sequential hypothesis pair:
\begin{equation}
\label{eq:hypothesis}
    \begin{aligned}
        H_0\!: &\quad \mathbb{E}_{P_t}[M_t \mid \mathcal{F}_{t-1}] \geq \epsilon \quad \forall t \in \{1, ..., T\} \\ 
        H_1\!: &\quad \exists t: \mathbb{E}_{P_t}[M_t \mid \mathcal{F}_{t-1}] < \epsilon. 
    \end{aligned}
\end{equation}
\looseness=-1 Here, $\mathbb{E}_{P_t}[M_t \mid \mathcal{F}_{t-1}]$ denotes the metric's expected value at time $t$, with $Y_t \sim P_t$ (and thus $M_t$) originating from time-dependent and possibly changing distributions. Here, $\mathcal{F}_{t-1}$ encapsulates all the historical information available up to time $t$, i.e. before $M_t$ is revealed\footnote{Formally, $\mathcal{F}_{t-1}$ is referred to as the filtration and means the generated $\sigma$-algebra as $\sigma(\{(Y_1, M_1), \dots, (Y_{t-1}, M_{t-1})\})$ \citep{e-book}.}. 
The null hypothesis $H_0$ in \autoref{eq:hypothesis} states that expected tracking quality will remain at or above tolerance $\epsilon$ across video frames, while the alternative $H_1$ identifies whether expected tracking quality drops below tolerance at any particular time during tracking, indicating what we term a \emph{tracking failure}.

\noindent To test the hypotheses, our goal is to employ a test statistic that efficiently and dynamically captures the metric's behaviour across video frames. In particular, we make use not only of the metric's current state, but inform our evidence for or against tracking failure via a sequential design that accumulates evidence over time. To this end, we employ an \emph{e-process}---a mathematical workhorse in the recently popularized testing-by-betting literature \citep{ramdas2023game}. Formally, an e-process $\{X_t\}_{t \ge 0}$ describes a non-negative process such that 
\begin{equation}
\label{eq:e-process-condition}
    \mathbb{E}_{H_0}{[X_t]} \leq 1
\end{equation}
for any arbitrary stopping time $t$, and mathematically forms a \emph{test supermartingale} (see \autoref{app:seq-testing}). \autoref{eq:e-process-condition} highlights that if $H_0$ is true (no failure) the value's expected growth is upper-bounded, regardless of when we choose to stop and make a test decision (termed `optional stopping'). In contrast, growth under $H_1$ has no upper limit, and can thus serve as a natural proxy to track violations of the metric against the selected tolerance level. Such an e-process derives its error-controlling qualities under $H_0$ via the application of the Ville's inequality \cite{ville1939etude}, guaranteeing that for any significance level $\alpha \in (0, 1)$ we have 
\begin{equation}
    \mathbb{P}_{H_0} \left( \sup_{t \ge 0} X_t \ge \frac{1}{\alpha} \right) \le \alpha,
    \label{eq:ville}
\end{equation}
that is, the probability of excessive growth beyond $\frac{1}{\alpha}$ across time steps is strictly bounded at rate $\alpha$. \autoref{eq:ville} in essence ensures dynamic and \emph{anytime-valid} Type-I error control, and naturally leads to the following practical test decision: track the growth of $X_t$ and reject $H_0$ as soon as its value exceeds $\frac{1}{\alpha}$. This is formalized by the \emph{stopping time} $\tau$, defined as
\begin{equation}
    \label{eq:e-process-tracking-failure}
    \tau = \inf \left\{ t: X_t \geq \frac{1}{\alpha}\right\}.
\end{equation}
\autoref{eq:ville} ensures that following this protocol erroneous rejections, \ie~mistakenly claiming tracking failures, occurs with at most probability $\alpha$ (where $\alpha \in (0, 0.1]$ is typically chosen, as is standard in the hypothesis testing literature \cite{fisher1970statistical}). A final question then is how $X_t$ is explicitly instantiated to grow quickly and ensure meaningful tracking signal. To that end, we leverage existing design principles from \citet{aGRAPA} and construct an e-process as
\begin{equation}
\label{eq:e-process}
    X_0 = 1, \; \;
    X_t = \prod_{i=1}^t  1 + \lambda_i (\epsilon - M_i).
\end{equation}
\looseness=-1 Starting from initial value $X_0 = 1$, $X_t$ multiplicatively incorporates tracking failure signals $(\epsilon - M_t)$ with rate $\lambda_t \in [0, 1]$, which is coined the \emph{betting rate} following a pre-specified update rule (see \autoref{subsec:method-bettingrate})\footnote{The term `betting rate' originates from game-theoretic interpretations of \autoref{eq:e-process} not further elaborated on, see \cite{ramdas2023game}.}. $\lambda_t$ strictly depends on past values $\{M_i\}_{i=1}^{t-1}$ (it is measurable w.r.t. $\mathcal{F}_{t-1}$), rendering it a predictable and dynamically adjustable quantity that may amplify or down-weigh evidence depending on past growth \cite{e-book}. When tracking quality is sufficient we have $(\epsilon - M_i) \leq 0$ and $1 + \lambda_i (\epsilon - M_i) \leq 1$, ensuring $X_t$ does not grow. When tracking fails we have $(\epsilon - M_i) > 0$ and $1 + \lambda_i (\epsilon - M_i) > 1$, ensuring fast growth of $X_t$ and quick accumulation of evidence against $H_0$. Combined with \autoref{eq:e-process-tracking-failure}, the sequential testing protocol is then clearly defined, and it remains to clarify how $M_t$ and $\lambda_t$ are instantiated. We defer further background and details on sequential testing to \autoref{sec:ap-e-processes} and \autoref{sec:ap-derivation}.

\subsection{Measuring Tracking Quality ($M_t$)}
\label{subsec:method-tracking-metrics}

\looseness=-1 The introduced sequential test is driven by the information fed through the tracking metric $M_t$. We next discuss both supervised and unsupervised instantiations of this metric, permitting flexible use of available information.

\paragraph{Supervised setting.} \looseness=-1 For this case, we assume access to the ground-truth bounding box $b_t$ for the object to be tracked, \eg~via human annotation or external validation. Thus, tracking quality can be optimally measured by comparison of this ground truth and the predicted bounding boxe $\hat{b}_t$ by leveraging standard metrics such as \emph{intersection-over-union} (IoU), and its generalized counterpart (GIoU). In particular, the GIoU \cite{giou} avoids degeneracy when no box overlap between $\hat{b}_t$ and $b_t$ exists, rendering it more informative as tracking quality degrades, and is computed as
\begin{equation}
\label{eq:giou}
    \text{GIoU}_t = \text{IoU}_t - \frac{|H  \; \backslash \; (\hat{b}_t \; \cup \; b_t)|}{|H|},
\end{equation}
with $H$ denoting the minimal convex hull (\ie~smallest rectangle) enclosing both bounding boxes. Since generally GIoU $\in [-1, 1]$, the metric is normalized as $\text{NGIoU}_t = \frac{\text{GIoU} + 1}{2}$ to adhere to the desired range in $[0, 1]$.

\paragraph{Unsupervised setting.} \looseness=-1 In this case no access to ground-truth annotations exists, and tracking quality thus needs to be derived from model-internal tracking signals, with tracking \emph{confidence} acting as a quality proxy. Trackers such as KCF and SiamFC derive predictions from internal response maps, capturing a notion of `target similarity' at different searched image locations. For instance, \citet{tracking-failure-KCF} leverage internal maps to analyse peak values across frames, informing tracking failures. In a similar fashion, a quality metric $M_t \in [0,1]$ can be defined that is amenable to our sequential test, and we consider some options below. 

\looseness=-1 We denote the model-internal response map at time $t$ as $C_t \in \mathbb{R}^{W' \times H'}$, where $W'$ and $H'$ form the dimensions of the searched image (which will differ from $W$ and $H$ due to internal feature reshaping), and a given value $C_t(i, j) \in [0,1]$ measures the similarity or correlation\footnote{Note this does not equate the statistical definition of correlation.} between target object and sub-frame centred at pixel location $(i, j)$. A straightforward measure of tracking confidence is then obtained by the map's \emph{peak correlation} (PC), \ie
\begin{equation}
\label{eq:peak-correlation}
    \text{PC}_t = \max_{(i, j) \in W' \times H'} C_t(i, j).
\end{equation}
\looseness=-1 A potential issue with peak correlation is that its magnitude can differ depending on the specific video, as some objects and videos are easier to track than others, which might imply higher correlation values. To address this, following \citet{tracking-failure-KCF} a normalized version of the metric coined \emph{certainty gain} (CG) is given by
\begin{equation}
\label{eq:certainty-gain}
    \text{CG}_t = \min \left\{ 1, \; \frac{\text{PC}_t}{\frac{1}{\sigma}(\sum_{i=t-\sigma+1}^t \text{PC}_i)}\right\},
\end{equation}
where $\sigma \leq t $ denotes a fixed window size determining the number of past frames used for normalization. Upper clipping by one is applied to ensure the target range. While informative, these two metrics only account for peak values and discard information about the remaining response map. Instead, information about the full distribution of correlations can be captured by the \emph{average peak-to-correlation energy} (APCE) \cite{object-tracking-apce}, defined as
\begin{equation}
\label{eq:apce}
    \text{APCE}_t = \frac{ \left| \text{PC}_t - \text{MC}_t \right|^2}{\frac{1}{|W' H'|} \left( \sum_{i=1}^{W'}  \sum_{j=1}^{H'}  \left(C_t(i, j) - \text{MC}_t\right)^2 \right)}.
\end{equation}
$\text{PC}_t$ denotes the peak correlation as before, while $\text{MC}_t = \min_{(i, j) \in W' \times H'} C_t(i, j)$ is the minimum correlation value. APCE effectively captures the entropy of the distribution---response maps with higher concentrations of large correlation yield a larger APCE, and vice versa. However, further normalization to $[0, 1]$ is required, and we thus follow the same structure as in \autoref{eq:certainty-gain} to obtain \emph{sharpness gain} (SG) as
\begin{equation}
\label{eq:sharpness-gain}
    \text{SG}_t=\min \left\{ 1, \; \frac{\text{APCE}_t}{\frac{1}{\sigma}(\sum_{i= t - \sigma + 1}^t \text{APCE}_i)} \right\},
\end{equation}
where once more $\sigma \leq t$ denotes a recency window.

\subsection{Learning the Betting Rate ($\lambda_t$)}
\label{subsec:method-bettingrate}

\looseness=-1 We want to ensure that our test design accumulates per-frame evidence as rapidly as possible should a true tracking failure occur, limiting the delay between occurrence and alert. This is ensured in parts by the multiplicative nature of \autoref{eq:e-process}, but also crucially driven by the choice of betting rate $\lambda_t$. As previously outlined, $\lambda_t$ may only depend on past values $\{M_i\}_{i=1}^{t-1}$, rendering it a predictable quantity and learnable from available data at every step $t$ \cite{e-book}. Reactive growth is formalized by the principle of \emph{growth-rate optimality} \cite{grunwald2024safe}, that is, we aim to select a betting rate $\lambda_t$ which maximizes the \emph{expected log-growth} of the process under tracking failure, or
\begin{equation}
\label{eq:opt-problem-betting-rate}
    \lambda_t = \arg\max_{\lambda \in [0, \frac{1}{2 \epsilon}]} \mathbb{E}_{H_1}[\log X_t \mid \mathcal{F}_{t-1}].
\end{equation}
Note that the initial set of possible choices is already restricted to $[0, \frac{1}{2\epsilon}]$ to ensure non-negativity and a valid e-process (see \autoref{app:seq-testing}). We next outline two considered options in this work, following recommendations in \citet{aGRAPA}.

\paragraph{Approximate GRAPA (aGRAPA).} \looseness=-1 This approach provides an approximate solution to the objective in \autoref{eq:opt-problem-betting-rate}, resulting in the efficient closed-form update rule for $\lambda_t$ as 
\begin{equation}
\label{eq:aGRAPA}
    \lambda_{t}^{\text{aGRAPA}} = \frac{\epsilon - \hat{\mu}_{t-1}}{\hat{\sigma}_{t-1}^2 + (\epsilon - \hat{\mu}_{t-1})^2}.
\end{equation}
$\hat{\mu}_{t-1}$ and $\hat{\sigma}_{t-1}^2$ form the empirical running mean and variance over $\{M_i\}_{i=1}^{t-1}$, and $\lambda_{t}^{\text{aGRAPA}}$ is subsequently clipped to $[0, \frac{1}{2 \epsilon}]$. Intuitively, the betting rate increases when the running mean is far from $\epsilon$, which is further amplified by a small variance.

\paragraph{Scale-Free Online Gradient Descent (SF-OGD).} \looseness=-1 Based on work in online learning by \citet{sf-ogd}, the second option updates $\lambda_t$ using the normalized gradient of the most recent observation's log-loss term $\ell(m_{t-1}; \lambda_{t-1}) = -\log(1 + \lambda_{t-1} (\epsilon - m_{t-1}))$, that is
\begin{equation}
\label{eq:sfogd}
    \lambda^{\text{SF-OGD}}_{t} = \lambda_{t-1} - \gamma \cdot \frac{\nabla_{\lambda_{t-1}} \ell(m_{t-1}; \lambda_{t-1})}{\sqrt{\sum_{i=1}^{t-1} (\nabla_{\lambda_i} \ell(M_i; \lambda_i))^2}},
\end{equation}
and likewise clipped to $\left[0, \frac{1}{2\epsilon}\right]$. $\gamma > 0$ constitutes a fixed learning rate, the gradient is given by $\nabla_{\lambda_{t-1}} \ell(m_{t-1}; \lambda_{t-1}) = - (\epsilon - m_{t-1})/(1 + \lambda_{t-1} (\epsilon - m_{t-1}))$, and normalization by past gradient norms ensures scale-invariance to gradient magnitudes. We refer to \citet{aGRAPA} for further discussion on suitable betting rate designs.

\section{Experimental Design}
\label{sec:experiments}

\looseness=-1 We next outline our experimental design, from datasets and tracking models to test parameters and evaluation metrics.

\begin{table}[!t]
\caption{
Comparison of empirically evaluated tracking datasets.
}
\label{tab:datasets}
\resizebox{\columnwidth}{!}{
\begin{tabular}{lccc}
    \toprule
    \textbf{Dataset} & \textbf{\# Videos} & \textbf{Avg. Duration} & \textbf{Frame Rate} \\
    \midrule
    OTB-100~\cite{otb2015}    & 100     & 20 s   & 30 FPS \\
    LaSOT~\cite{lasot}      & 1400     & 84 s   & 30 FPS \\
    TrackingNet~\cite{trackingnet} & 31k     & 16 s   & 30 FPS \\
    GOT-10k~\cite{got-10k}    & 10k      & 15 s   & 10 FPS \\
    \bottomrule
\end{tabular}
}
\end{table}

\paragraph{Datasets and tracking models.} \looseness=-1 We employ four widely used object tracking benchmarks shown in \autoref{tab:datasets}, and evaluate on $N_{\text{vid}}=50$ randomly sampled videos for each one. These datasets collectively span a wide range of tracking conditions that include occlusions, fast motion, illumination changes, and scale variation. They also include varying video lengths and motion speeds, such as longer and slower-paced videos in LaSOT \emph{vs.} short and fast videos in GOT-10k. Collectively, our choices ensure robust evaluation across multiple challenging tracking settings. We consider KCF \cite{kcf} and SiamFC \cite{siamfc} as our two different tracking models, noting that the testing framework is in principle \emph{model-agnostic} and amenable to any underlying tracker, permitting proper access to validate $M_t$ when obtained from model-internal signals. KCF is available in OpenCV \cite{bradski2000opencv} with minor adjustments to properly access response maps, while SiamFC is implemented from a public repository\footnote{\url{https://github.com/huanglianghua/siamfc-pytorch}}. 

\paragraph{Recency and window size.} \looseness=-1 Our e-process is consistently built following \autoref{eq:e-process-tracking-failure}, and combined with either aGRAPA or SF-OGD betting rate strategies and any of the tracking quality metrics outlined in \autoref{tab:thresholds}. To mitigate abrupt peaks and variability in the metric and stabilize learning we employ a smoothing exponential moving average (factor $0.25$) across values. This does not violate testing design, as we may leverage past information any way we see fit \cite{aGRAPA}. Indeed, basing the betting rates on all past values can hamper reactiveness, as older information (\eg~early frames with acceptable tracking) may not be as relevant anymore for the current state. Thus we also restrict the betting rates to a fixed recency window of $w_\text{E}$ past frames, set to equate two seconds for the supervised setting (\eg~$w_\text{E} = 20$ for GOT-10k) and fixed $w_\text{E} = 10$ for the unsupervised case. The same window size ($\sigma=10$) is used in \autoref{eq:certainty-gain} and \autoref{eq:sharpness-gain}.

\paragraph{Tolerance and significance level.} \looseness=-1 \autoref{eq:hypothesis} relies on fixed, user-specified parameters $\epsilon$ and $\alpha$ to determine when tracking is deemed insufficient, and when a final alert should be raised. The tolerance level $\epsilon$ acts as a key reference point subduing or driving e-process growth, and we opt for different fixed values according to each metric in \autoref{tab:thresholds}. These are set based on inspecting general metric behaviour and value ranges, but users are free to choose according to their particular application and requirements. Specifically, since GIoU and peak correlation are absolute measures of tracking quality we set their thresholds near the range midpoints. In contrast, certainty gain and sharpness gain are relative to previous timesteps, thus thresholds closer to one are more suitable. Finally, we evaluate at fixed significance level $\alpha=0.1$, as is common in the testing literature \cite{casella2002statistical}. Subsequently, our rejection threshold for testing is $\frac{1}{\alpha} = 10$.

\begin{table}[!t]
    \caption{
    Fixed tolerance levels $\epsilon$ across considered quality metrics.
    }
    \label{tab:thresholds}
    \centering
    \resizebox{0.85\columnwidth}{!}{
    \begin{tabular}{lc}
    	\toprule
    	\textbf{Tracking quality metric} $M_t$ & \textbf{Tolerance level} $\epsilon$ \\
    	\midrule
    	Normalized GIoU ($\text{NGIoU}_t$) & $0.55$ \\ 
        Peak Correlation ($\text{PC}_t$) & $0.50$ \\ 
        Certainty Gain ($\text{CG}_t$) & $0.95$ \\ 
        Sharpness Gain ($\text{SG}_t$) & $0.90$ \\
        \bottomrule
    \end{tabular}
    }
\end{table}

\paragraph{Evaluation metrics.} \looseness=-1 To assess utility and correctness of raised tracking failure alerts, empirical stopping times $\hat{\tau}$ following \autoref{eq:e-process-tracking-failure} need to be compared to a notion of ground-truth failure. As true tracking failure signals are generally not provided with object tracking datasets, we construct a heuristic ground truth defined by the first time step for which $M_t < \epsilon$ holds for a user-defined number of $w_\text{GT}$ frames\footnote{If never evaluated true for a particular video, we set $w_\text{GT} = T$.}. That is, true tracking failure is associated with robust quality degradation, where lower $w_\text{GT}$ generally provides more reactive but volatile signals, and vice versa. In line with fixed recency windows, $w_\text{GT}$ is set to equate two seconds for the supervised setting, and $w_\text{GT}=10$ frames for the unsupervised case. A window of two seconds (translated to frames by \autoref{tab:datasets}) provides reasonable temporal buffer to confirm tracking failure beyond quick recovery, balancing tolerance for brief errors with the need to flag sustained failures. For unsupervised metrics a generally shorter window is more appropriate, as the tracker may lock onto different objects (which are not distinguishable bar ground-truth bounding boxes), causing perceived tracking recovery despite actual failure. A smaller window increases robustness to such rapid but incorrect reassignments. 


\looseness=-1 Proper empirical measurement of obtained false alert assurances (\autoref{eq:ville}) at level $\alpha$ necessitates experiment repetition across multiple instantiations of the test, since guarantees hold individually for each test associated with a single tracking video. Therefore mild random Gaussian noise is injected to simulate $N_{\text{trials}}=50$ randomized repetitions of each video alongside the original. Given thus defined ground-truth failure time $\tau$, empirical stopping time $\hat{\tau}$, and randomization, our two key desiderata to capture are validity and speed of raised alerts, measured as follows:

\looseness=-1 \emph{False Positive Rate} (FPR). The empirical fraction of false alerts, \ie~prematurely claiming tracking failure and thus cases where $\hat{\tau} < \tau$ holds true. Results are averaged across both different video samples and per-video trials, yielding the final aggregate
\begin{equation}
\label{eq:fpr}
    \text{FPR} = \frac{1}{N_{\text{vid}} \cdot N_{\text{trials}}} \sum_{i=1}^{N_{\text{vid}}} \sum_{j=1}^{N_{\text{trials}}} \mathbf{1}[\hat{\tau}^{(i, j)} < \tau^{(i)}],
\end{equation}
with indicator function $\mathbf{1}[\cdot]$. Validity is maintained when $\text{FPR} \leq \alpha$, \ie~theoretical and empirical assurances align.

\looseness=-1 \emph{Average Detection Delay} (ADD). The average gap between true and raised stopping times, that is 
\begin{equation}
\label{eq:add}
\text{ADD} = \frac{1}{N_{\text{vid}} \cdot N_{\text{trials}}} \sum_{i=1}^{N_{\text{vid}}} \sum_{j=1}^{N_{\text{trials}}} \left( \hat \tau^{(i,j)} - \tau^{(i)} \right).
\end{equation}
Since false alerts result in $(\hat \tau^{(i,j)} - \tau^{(i)}) < 0$, we generally compute the metric only for true positives or correctly identified failures, thus measuring the average \emph{detection delay} desired to be as small as possible.
\section{Experimental Results}
\label{sec:results}

\begin{figure}[t]
  \centering
  \includegraphics[width=\linewidth]{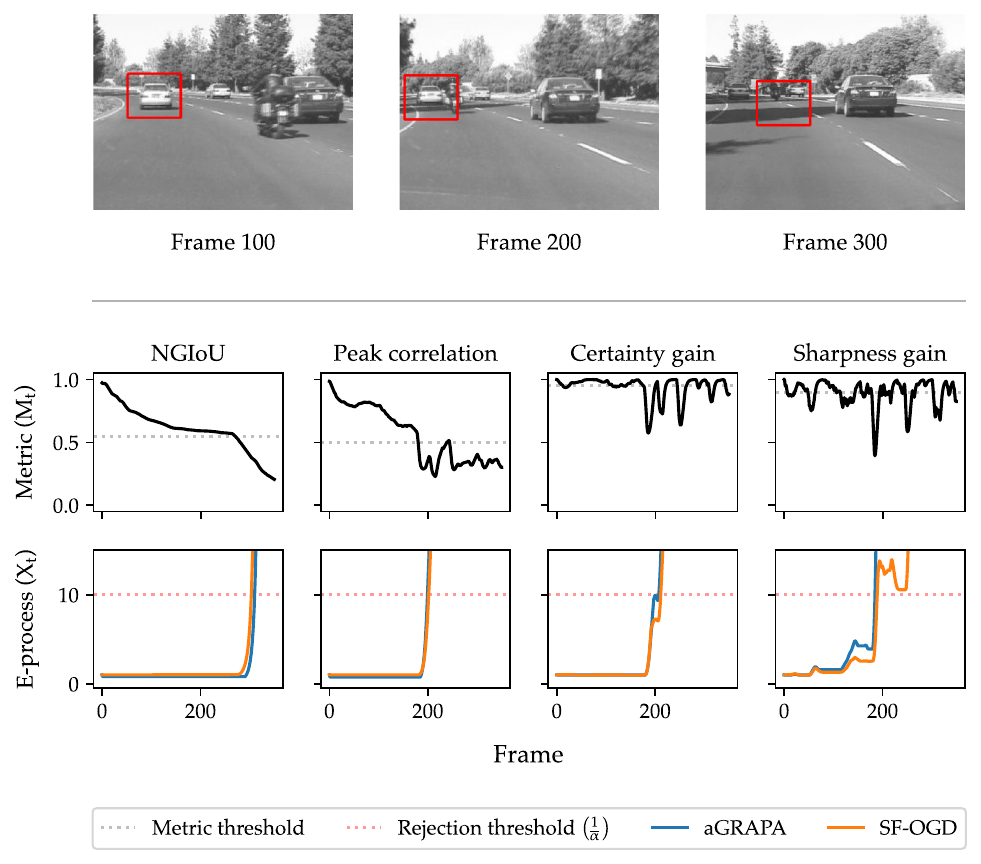}
  \caption{
  An example of effective tracking failure detection for different tracking quality metrics $M_t$ (\emph{top}) and corresponding e-processes $X_t$ (\emph{bottom}) leveraging both aGRAPA and SF-OGD betting rates (\autoref{subsec:method-bettingrate}). The video sample (\emph{Car-1}, OTB-100) depicts vehicle tracking in a traffic sequence, with subsequent tracking drift caused by a motorcycle occlusion. Failure alerts are raised consistently across metrics, with supervised (NGIoU) reacting later but more stably, while unsupervised ones tend to more volatility.
  }
  \label{fig:example1}
\end{figure}

\begin{figure}[!h]
  \centering
  \includegraphics[width=\linewidth]{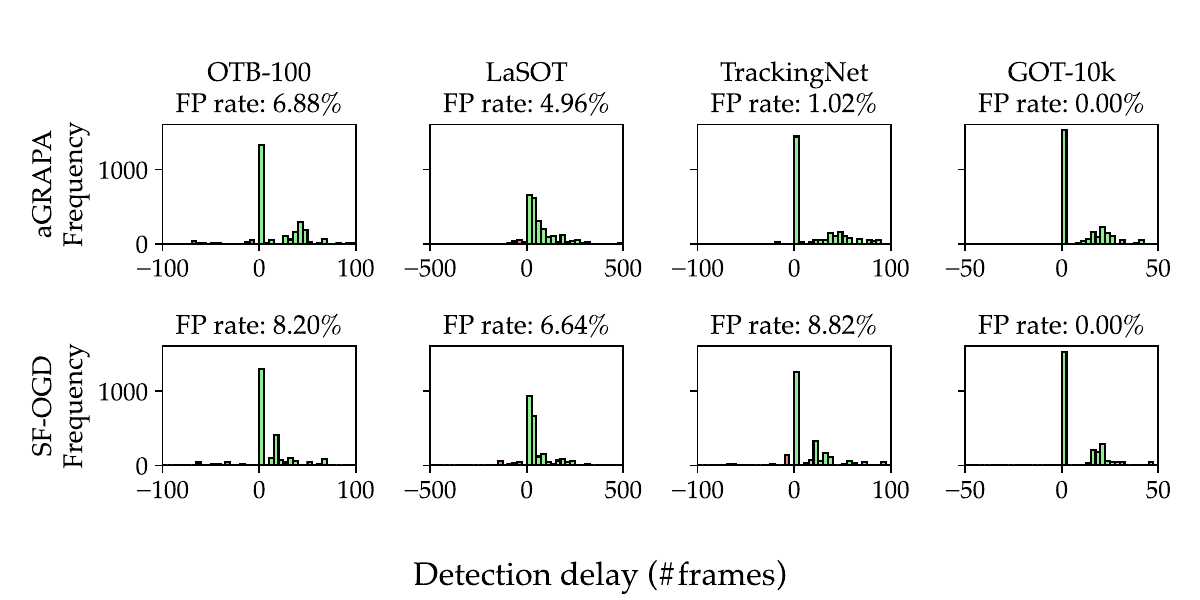}
  \caption{
  Histograms depicting the distribution of detection delays across datasets and both aGRAPA and SF-OGD betting rates (\autoref{subsec:method-bettingrate}), for an e-process with supervised NGIoU metric (\autoref{subsec:method-tracking-metrics}). False positives (premature alerts) are marked red and denote the small and contained fraction of negative detection delays. Overall, detection delays are relatively small and concentrate close to zero, indicating quick and reactive failure alerts by sequential testing.
  }
  \label{fig:delay_hist}
\end{figure}

\begin{figure}[!h]
	\includegraphics[width=\linewidth]{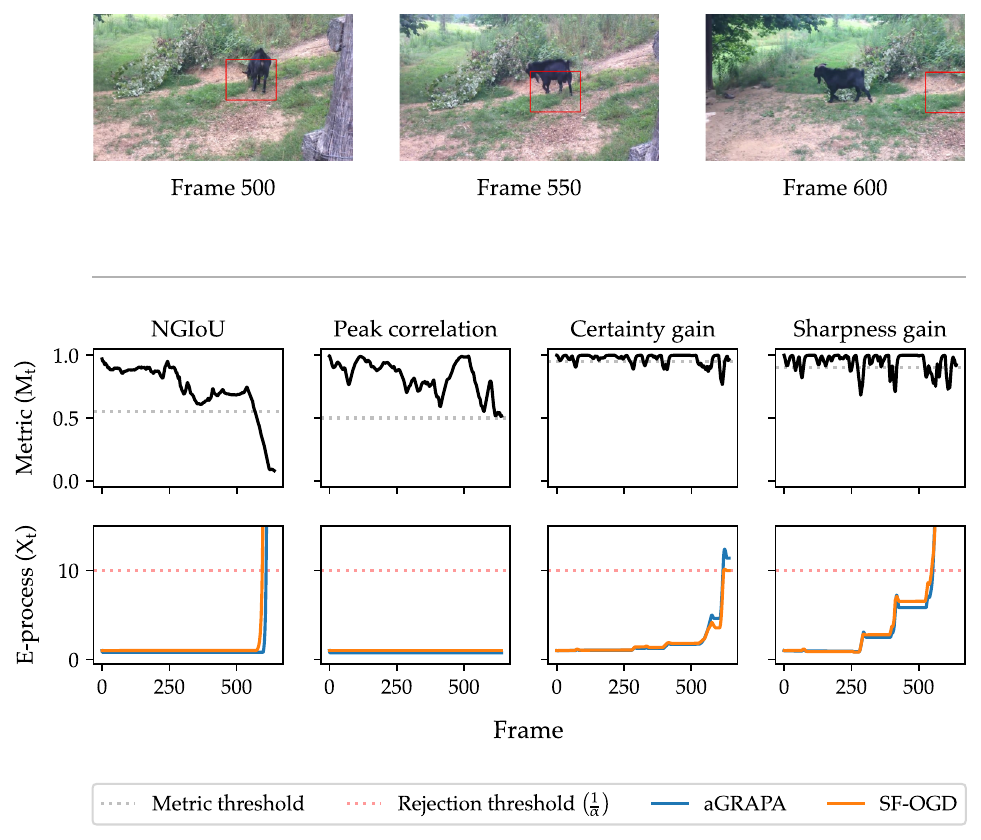}
	\caption{
    An example of partially unsuccessful tracking failure detection, for different tracking quality metrics $M_t$ (\emph{top}) and corresponding e-processes $X_t$ (\emph{bottom}) leveraging both aGRAPA and SF-OGD betting rates (\autoref{subsec:method-bettingrate}). The video sample (\emph{Sheep-1}, LaSOT) depicts animal tracking in a natural environment, with subsequent tracking drift caused by rapid object movements. While clearly identified by the supervised signal (NGIoU), unsupervised metrics struggle more with raising a proper alert, peak correlation failing to do so entirely.
    }
    \label{fig:qualitative-sheep}
\end{figure}

\begin{figure}[t]
	\includegraphics[width=\linewidth]{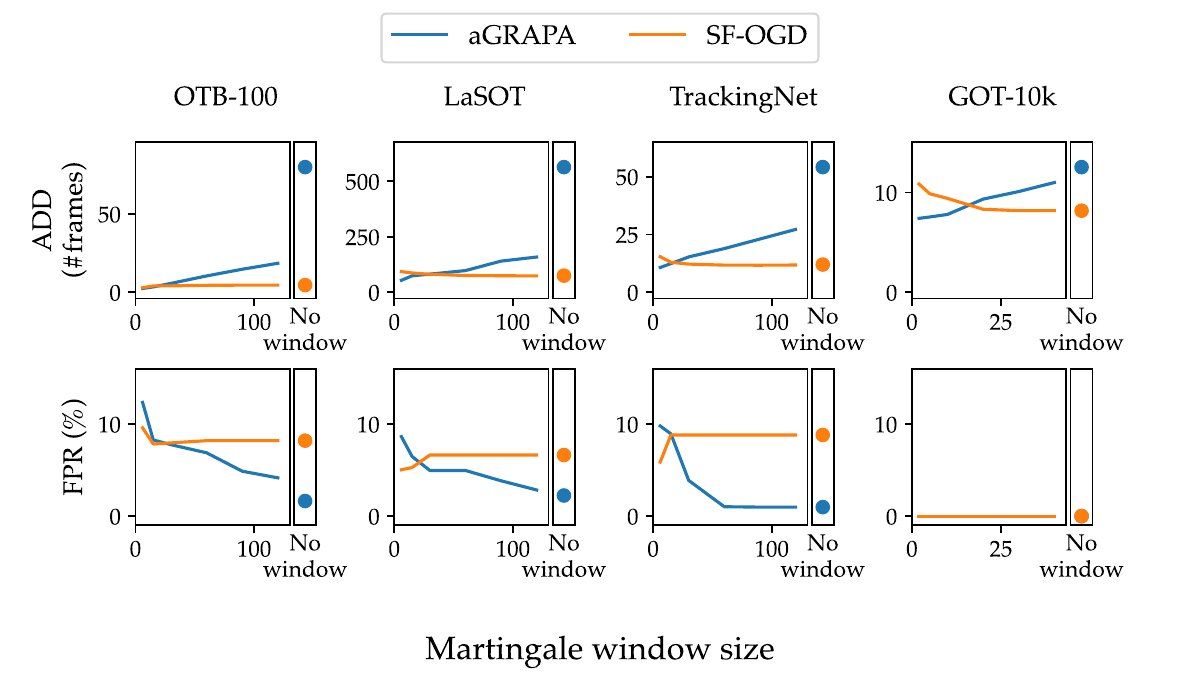}
	\caption{
    Ablation study on the effects of selecting the recency window size $w_{\text{E}}$ across datasets, betting rates, and metrics (\autoref{sec:experiments}). $w_{\text{E}}$ effectively truncates past frames to yield more reactive e-process growth, as seen by lower detection delays for aGRAPA with small windows. On the other hand, validity is ensured regardless, as seen by consistently contained FPR. `No window' refers to no truncation, employing the full tracking history per video.
    }
    \label{fig:supervised-delay-window}
\end{figure}

We next describe and discuss our experimental results, employing a mix of qualitative examples (\autoref{fig:example1}, \autoref{fig:qualitative-sheep}) and aggregate metrics (\autoref{tab:quantitative}, \autoref{fig:delay_hist}, \autoref{fig:supervised-delay-window}).

\paragraph{Evidence accumulation.} \autoref{fig:example1} illustrates the fundamental behaviour of our proposed sequential testing framework with a representative example. As the video progresses and the tracker starts to drift and lose its target, degradation in tracking quality is captured by significant drops in metric values, either directly via a supervised signal (NGIoU) or via proxy by reduced tracking confidence for unsupervised metrics. Once fallen below their fixed tolerance levels $\epsilon$ (\autoref{tab:thresholds}), failure signals accumulate to rapidly grow the e-process, triggering an alert once $\frac{1}{\alpha}=10$ is crossed. In contrast, e-processes remain low and stable as long as tracking is deemed sufficiently reliable. We stress how unsupervised signals are more volatile and may increase again, since they do not disambiguate between effective tracking of different objects. This motivated our choice of shorter ground-truth frame count $w_{\text{GT}}$, and additional measures could be explored to properly distinguish when drops and recoveries are positive, and when they are misleading.

\begin{table*}[h!]
\centering
\caption{
Aggregated tracking failure detection results across two tracking models (KCF, SiamFC) and all four video datasets. We consider both aGRAPA and SF-OGD betting rates (\autoref{subsec:method-bettingrate}), for an e-process with supervised NGIoU metric (\autoref{subsec:method-tracking-metrics}). Evaluation metrics FPR (\autoref{eq:fpr}) and ADD (\autoref{eq:add}) are reported with mean (and stdev.) across $N_{\text{vid}}=50$ video samples per dataset, and $N_{\text{trials}}=50$ randomizations.
}
\label{tab:quantitative}
\resizebox{\textwidth}{!}{
\begin{tabular}{llcccccccc}
\toprule
\textbf{Tracker} & \textbf{Bets} $\lambda_t$ & \multicolumn{2}{c}{\textbf{OTB-100}} & \multicolumn{2}{c}{\textbf{LaSOT}} & \multicolumn{2}{c}{\textbf{TrackingNet}} & \multicolumn{2}{c}{\textbf{GOT-10k}} \\
\cmidrule(lr){3-4} \cmidrule(lr){5-6} \cmidrule(lr){7-8} \cmidrule(lr){9-10} 
 & & FPR & ADD & FPR & ADD & FPR & ADD & FPR & ADD \\
\midrule
\multirow{2}{*}{\textbf{KCF}} 
 & aGRAPA & 6.88\% & 43.90 $\pm$ 4.79 & 4.96\% & 122.37 $\pm$ 15.27 & 1.02\% & 48.62 $\pm$ 2.39 & 0.00\% & 22.47 $\pm$ 0.41 \\
 & SF-OGD & 8.20\% & 31.26 $\pm$ 2.51 & 6.64\% & 102.13 $\pm$ 23.82 & 8.82\% & 34.73 $\pm$ 1.69 & 0.00\% & 21.44 $\pm$ 0.42 \\
\midrule
\multirow{2}{*}{\textbf{SiamFC}} 
 & aGRAPA & 6.12\% & 44.93 $\pm$ 1.63 & 0.60\% & 177.62 $\pm$ 81.42 & 0.00\% & 56.79 $\pm$ 3.31 & 0.00\% & 27.77 $\pm$ 1.88 \\
 & SF-OGD & 6.53\% & 25.65 $\pm$ 0.73 & 9.60\% & 104.97 $\pm$ 53.54 & 5.45\% & 39.37 $\pm$ 1.75 & 0.00\% & 26.15 $\pm$ 2.17 \\
\bottomrule
\end{tabular}
}
\end{table*}

\paragraph{Statistical validity and detection delays.} Aggregated results examining validity of the approach and mean detection delays are summarized in \autoref{tab:quantitative}, for various combinations of tracking model, dataset, and betting strategy. We opt to focus on an e-process with supervised NGIoU metric, as the constructed ground-truth failure signal ($M_t < \epsilon$ for $w_{\text{GT}}$ frames) employs supplied object bounding boxes and thus roots itself most reliably in the actual data. We observe that $\text{FPR} \leq 10\%$ holds throughout tests, empirically validating theoretical assurances provided by \autoref{eq:ville}. Indeed, rates are at times substantially more conservative (even achieving $\text{FPR}=0\%$ for GOT-10k), implying that further trade-offs between false positives and reduction in delays are possible, primarily by more involved design choices targeting yet faster reaction. While average detection delays vary across combinations of model, dataset, and betting strategy, they are generally low, indicating good responsiveness. This is supported by examining the full empirical distributions more closely in \autoref{fig:delay_hist}, where delays cluster close to zero and are skewed by a few notable large delays. The magnitude of delays is visibly affected by the choice of dataset, which we attribute in part to perceived evaluation challenges. In particular, tracked objects in the LaSOT dataset tend to move at slower speeds, incurring a more gradual metric degradation and slower e-process growth which can delay detection. 

\paragraph{Choice of tracking metrics and thresholds.} The e-process in \autoref{eq:e-process} primarily parses tracking failure signals through the defined quality metric $M_t$, and to a lesser extent via the rate $\lambda_t$. Thus the key design choice relates to an accurate reflection of per-frame tracking quality. While the NGIoU is reliable due its explicit rooting in the data, its supervised nature makes it less suitable for low-latency, real-world tracking scenarios, hence we explore alternative unsupervised notions employing confidence proxies. Yet, \autoref{fig:qualitative-sheep} clearly outlines that not all metrics are equally stable, or sensitive to failure cases. In the particular example, peak correlation remains high despite tracking drift, resulting in zero e-process growth and a missed alert. In contrast, certainty gain and sharpness gain record multiple confidence drops, eventually accumulating sufficient evidence for meaningful growth. Yet, it is clear that timely alerts are highly dependent on the choice of $M_t$, and its relation to the tolerance level $\epsilon$. Overly loose thresholds may substantially delay detection, while highly conservative ones heighten the risk of false alarms (yet remain contained at rate $\alpha$). Without invalidating theoretical assurances provided by the framework, its \emph{practical} use requires careful consideration of the e-process' design parameters and their trade-offs.

\paragraph{Recency window size and betting rates.} An effective and practically-oriented design choice is the fixed-size recency window $w_{\text{E}}$, truncating past metric values deemed uninformative for the current tracking state when updating $\lambda_t$. We ablate against varying window sizes in \autoref{fig:supervised-delay-window}, from extremely short windows to no truncation at all (`No window'). As expected, larger window sizes provide stability by smoothing out short-term fluctuations but can also delay detection in fast-changing scenarios, as running quantities exhibit higher `inertia' and down-weigh immediate evidence. The effect is particularly pronounced for longer videos such as those found in LaSOT, where delays increase by a noticeable number of frames. When comparing aGRAPA and SF-OGD, the latter appears less sensitive to these effects, suggesting the gradient signal is sufficiently strong to counteract window-dependent normalization. Indeed, the impact on running mean and variance terms in \autoref{eq:aGRAPA} is clearly more immediate. Despite impacts on detection delay, validity remains less affected and with small fluctuations is positively contained at target rate $\alpha$. 

Connecting back to results in \autoref{tab:quantitative}, the betting rate design of aGRAPA tends to be slightly more conservative overall, with lower false alerts but larger delays, whereas SF-OGD is able to react more adaptively to sudden tracking changes despite low sensitivity to $w_{\text{E}}$. This is in alignment with how each strategy is defined, as aGRAPA considers each evidence term equally in its running quantities, while SF-OGD is primarily guided by the gradient direction derived from its latest evidence signal. 

\section{Conclusion}
\label{sec:conclusion}

\looseness=-1 We present a statistically grounded framework for real-time detection of object tracking failures by interpreting online tracking as a sequential hypothesis testing problem. Leveraging the theory of e-processes, our test provides anytime-valid control of false alerts while rapidly accumulating evidence once true failures emerge. The resulting mechanism is lightweight, model-agnostic, and operates in both supervised and unsupervised configurations, requiring only externally provided ground truth or readily available tracker-internal signals. Across two representative tracker families and four video benchmarks, we observe consistently contained false positive rates and low detection delays, demonstrating that sequential testing can act as an effective monitoring layer for deployed tracking systems.

\noindent A key advantage of the approach is its flexibility, permitting easy integration with existing pipelines and no modifications to the tracking algorithm or model itself, while imposing negligible computational overhead. Nonetheless, its performance ultimately depends on the informativeness of the underlying tracking quality metric. In cases where the metric degrades slowly or remains stable despite drift, detections may be delayed or missed altogether. Thus the use of more informative unsupervised metrics or internal state tracking mechanisms should lead to further performance gains. Moreover, while our experiments incorporate multiple datasets and trackers, additional work is required to stress-test performance under more elaborate and dynamic environments involving multiple occlusions, target-switching scenarios, camouflaging, anomalies, and others. 

\paragraph{Future work.} \looseness=-1 Several avenues remain open for future research. A natural next step is to couple failure detection with automated tracker adaptation or re-initialization \cite{tracking-failure-KCF}, enabling not only failure detection and monitoring but also recovery. Exploring richer or learned tracking-quality signals may further strengthen detection in challenging regimes, as motivated above. Finally, the broader methodology of sequential testing with e-processes is well suited to other streaming vision tasks, including multi-object tracking \cite{luo2021multiple}, action recognition \cite{action-rec-survey}, trajectory prediction \cite{rudenko2020human}, or anomaly detection \cite{ramachandra2020survey}, for which statistically principled failure detection is equally beneficial. Ultimately, equipping online tracking pipelines with principled failure detection offers a step towards systems that can monitor themselves, recognize when they are no longer reliable, and adapt safely in real time.

\clearpage

{
    \small
    \bibliographystyle{ieeenat_fullname}
    \bibliography{main}

\begin{thebibliography}{56}
\providecommand{\natexlab}[1]{#1}
\providecommand{\url}[1]{\texttt{#1}}
\expandafter\ifx\csname urlstyle\endcsname\relax
  \providecommand{\doi}[1]{doi: #1}\else
  \providecommand{\doi}{doi: \begingroup \urlstyle{rm}\Url}\fi

\bibitem[Bar et~al.(2025)Bar, Shaer, and Romano]{entropy-matching}
Yarin Bar, Shalev Shaer, and Yaniv Romano.
\newblock Protected test-time adaptation via online entropy matching: A betting approach.
\newblock \emph{Advances in Neural Information Processing Systems}, 2025.

\bibitem[Bertinetto et~al.(2016)Bertinetto, Valmadre, Henriques, Vedaldi, and Torr]{siamfc}
Luca Bertinetto, Jack Valmadre, Joao~F Henriques, Andrea Vedaldi, and Philip~HS Torr.
\newblock Fully-convolutional siamese networks for object tracking.
\newblock \emph{European Conference on Computer Vision}, 2016.

\bibitem[Bhat et~al.(2018)Bhat, Johnander, Danelljan, Khan, and Felsberg]{bhat2018unveiling}
Goutam Bhat, Joakim Johnander, Martin Danelljan, Fahad~Shahbaz Khan, and Michael Felsberg.
\newblock Unveiling the power of deep tracking.
\newblock \emph{European Conference on Computer Vision}, 2018.

\bibitem[Bradski(2000)]{bradski2000opencv}
Gary Bradski.
\newblock The opencv library.
\newblock \emph{Dr. Dobb's Journal: Software Tools for the Professional Programmer}, 2000.

\bibitem[Casella and Berger(2002)]{casella2002statistical}
George Casella and Roger~L Berger.
\newblock \emph{Statistical inference}.
\newblock Duxbury Pacific Grove, CA, 2002.

\bibitem[Chandrakar et~al.(2022)Chandrakar, Raja, Miri, Sinha, Kushwaha, and Raja]{traffic-monitoring}
Ramakant Chandrakar, Rohit Raja, Rohit Miri, Upasana Sinha, Alok Kumar~Singh Kushwaha, and Hiral Raja.
\newblock Enhanced the moving object detection and object tracking for traffic surveillance using rbf-fdlnn and cbf algorithm.
\newblock \emph{Expert Systems with Applications}, 2022.

\bibitem[Chen et~al.(2022)Chen, Wang, Zhao, Lv, and Niu]{object-tracking-survey}
Fei Chen, Xiaodong Wang, Yunxiang Zhao, Shaohe Lv, and Xin Niu.
\newblock Visual object tracking: A survey.
\newblock \emph{Computer Vision and Image Understanding}, 2022.

\bibitem[Chen et~al.(2021)Chen, Yan, Zhu, Wang, Yang, and Lu]{chen2021transformer}
Xin Chen, Bin Yan, Jiawen Zhu, Dong Wang, Xiaoyun Yang, and Huchuan Lu.
\newblock Transformer tracking.
\newblock \emph{Conference on Computer Vision and Pattern Recognition}, 2021.

\bibitem[Chicco(2021)]{chicco2021siamese}
Davide Chicco.
\newblock Siamese neural networks: An overview.
\newblock \emph{Artificial Neural Networks}, 2021.

\bibitem[Danelljan et~al.(2019)Danelljan, Bhat, Khan, and Felsberg]{danelljan2019atom}
Martin Danelljan, Goutam Bhat, Fahad~Shahbaz Khan, and Michael Felsberg.
\newblock Atom: Accurate tracking by overlap maximization.
\newblock 2019.

\bibitem[Demirbas(2007)]{demirbas2007maneuvering}
Kerim Demirbas.
\newblock Maneuvering target tracking with hypothesis testing.
\newblock \emph{IEEE Transactions on Aerospace and Electronic Systems}, 2007.

\bibitem[Enescu et~al.(2007)Enescu, Ravyse, and Sahli]{enescu2007visual}
Valentin Enescu, Ilse Ravyse, and Hichem Sahli.
\newblock Visual tracking by hypothesis testing.
\newblock 2007.

\bibitem[Fan et~al.(2019)Fan, Lin, Yang, Chu, Deng, Yu, Bai, Xu, Liao, and Ling]{lasot}
Heng Fan, Liting Lin, Fan Yang, Peng Chu, Ge Deng, Sijia Yu, Hexin Bai, Yong Xu, Chunyuan Liao, and Haibin Ling.
\newblock Lasot: A high-quality benchmark for large-scale single object tracking.
\newblock \emph{Conference on Computer Vision and Pattern Recognition}, 2019.

\bibitem[Fisher(1970)]{fisher1970statistical}
Ronald~Aylmer Fisher.
\newblock Statistical methods for research workers.
\newblock \emph{Breakthroughs in statistics: Methodology and distribution}, 1970.

\bibitem[Gr{\"u}nwald et~al.(2024)Gr{\"u}nwald, Heide, and Koolen]{grunwald2024safe}
Peter Gr{\"u}nwald, Rianne~de Heide, and Wouter Koolen.
\newblock Safe testing.
\newblock \emph{Journal of the Royal Statistical Society B}, 2024.

\bibitem[Guo et~al.(2022)Guo, Wang, Yang, Wang, Zhang, Guo, Gao, and Guo]{autonomous-driving}
Shuman Guo, Shichang Wang, Zhenzhong Yang, Lijun Wang, Huawei Zhang, Pengyan Guo, Yuguo Gao, and Junkai Guo.
\newblock A review of deep learning-based visual multi-object tracking algorithms for autonomous driving.
\newblock \emph{Applied Sciences}, 2022.

\bibitem[Henriques et~al.(2014)Henriques, Caseiro, Martins, and Batista]{kcf}
Jo{\~{a}}o~F. Henriques, Rui Caseiro, Pedro Martins, and Jorge Batista.
\newblock High-speed tracking with kernelized correlation filters.
\newblock \emph{IEEE Transactions on Pattern Analysis and Machine Intelligence}, 2014.

\bibitem[Huang et~al.(2019)Huang, Zhao, and Huang]{got-10k}
Lianghua Huang, Xin Zhao, and Kaiqi Huang.
\newblock Got-10k: A large high-diversity benchmark for generic object tracking in the wild.
\newblock \emph{IEEE Transactions on Pattern Analysis and Machine Intelligence}, 2019.

\bibitem[KC et~al.(2012)KC, Delannay, Jacques, and De~Vleeschouwer]{kc2012iterative}
Amit~Kumar KC, Damien Delannay, Laurent Jacques, and Christophe De~Vleeschouwer.
\newblock Iterative hypothesis testing for multi-object tracking with noisy/missing appearance features.
\newblock \emph{Asian Conference on Computer Vision}, 2012.

\bibitem[Knyaz et~al.(2016)Knyaz, Zheltov, and Vishnyakov]{motion-analysis}
Vladimir~A Knyaz, Sergey~Yu Zheltov, and Boris~V Vishnyakov.
\newblock Robust object tracking techniques for vision-based 3d motion analysis applications.
\newblock \emph{Optics, Photonics and Digital Technologies for Imaging Applications}, 2016.

\bibitem[Kong and Fu(2022)]{action-rec-survey}
Yu Kong and Yun Fu.
\newblock Human action recognition and prediction: {A} survey.
\newblock \emph{International Journal of Computer Vision}, 2022.

\bibitem[Kumar et~al.(2024)Kumar, Vohra, Jain, Li, Gan, and Jain]{kumar2024correlation}
Ashish Kumar, Rubeena Vohra, Rachna Jain, Muyu Li, Chenquan Gan, and Deepak~Kumar Jain.
\newblock Correlation filter based single object tracking: A review.
\newblock \emph{Information Fusion}, 2024.

\bibitem[Lee and Waslander(2024)]{lee2024uncertainty}
Chang~Won Lee and Steven~L. Waslander.
\newblock Uncertaintytrack: Exploiting detection and localization uncertainty in multi-object tracking.
\newblock \emph{IEEE International Conference on Robotics and Automation}, 2024.

\bibitem[Li et~al.(2018)Li, Yan, Wu, Zhu, and Hu]{li2018high}
Bo Li, Junjie Yan, Wei Wu, Zheng Zhu, and Xiaolin Hu.
\newblock High performance visual tracking with siamese region proposal network.
\newblock \emph{Conference on Computer Vision and Pattern Recognition}, 2018.

\bibitem[Lukezic et~al.(2017)Lukezic, Vojir, ˇCehovin~Zajc, Matas, and Kristan]{lukezic2017discriminative}
Alan Lukezic, Tomas Vojir, Luka ˇCehovin~Zajc, Jiri Matas, and Matej Kristan.
\newblock Discriminative correlation filter with channel and spatial reliability.
\newblock \emph{Conference on Computer Vision and Pattern Recognition}, 2017.

\bibitem[Luo et~al.(2021)Luo, Xing, Milan, Zhang, Liu, and Kim]{luo2021multiple}
Wenhan Luo, Junliang Xing, Anton Milan, Xiaoqin Zhang, Wei Liu, and Tae-Kyun Kim.
\newblock Multiple object tracking: A literature review.
\newblock \emph{Artificial Intelligence}, 2021.

\bibitem[Ma et~al.(2015)Ma, Yang, Zhang, and Yang]{ma2015long}
Chao Ma, Xiaokang Yang, Chongyang Zhang, and Ming-Hsuan Yang.
\newblock Long-term correlation tracking, 2015.

\bibitem[M{\"{u}}ller et~al.(2018)M{\"{u}}ller, Bibi, Giancola, Al{-}Subaihi, and Ghanem]{trackingnet}
Matthias M{\"{u}}ller, Adel Bibi, Silvio Giancola, Salman Al{-}Subaihi, and Bernard Ghanem.
\newblock Trackingnet: {A} large-scale dataset and benchmark for object tracking in the wild.
\newblock \emph{European Conference on Computer Vision}, 2018.

\bibitem[Ojha and Sakhare(2015)]{video-surveillance}
Shipra Ojha and Sachin Sakhare.
\newblock Image processing techniques for object tracking in video surveillance- a survey.
\newblock \emph{International Conference on Pervasive Computing}, 2015.

\bibitem[Orabona and P{\'a}l(2018)]{sf-ogd}
Francesco Orabona and D{\'a}vid P{\'a}l.
\newblock Scale-free online learning.
\newblock \emph{Theoretical Computer Science}, 2018.

\bibitem[Podkopaev and Ramdas(2022)]{podkopaev2021tracking}
Aleksandr Podkopaev and Aaditya Ramdas.
\newblock Tracking the risk of a deployed model and detecting harmful distribution shifts.
\newblock \emph{International Conference on Learning Representations}, 2022.

\bibitem[Ramachandra et~al.(2020)Ramachandra, Jones, and Vatsavai]{ramachandra2020survey}
Bharathkumar Ramachandra, Michael~J Jones, and Ranga~Raju Vatsavai.
\newblock A survey of single-scene video anomaly detection.
\newblock \emph{IEEE Transactions on Pattern Analysis and Machine Intelligence}, 2020.

\bibitem[Ramdas and Wang(2025)]{e-book}
Aaditya Ramdas and Ruodu Wang.
\newblock \emph{Hypothesis testing with e-values}.
\newblock Foundations and Trends in Statistics, 2025.

\bibitem[Ramdas et~al.(2023)Ramdas, Gr{\"u}nwald, Vovk, and Shafer]{ramdas2023game}
Aaditya Ramdas, Peter Gr{\"u}nwald, Vladimir Vovk, and Glenn Shafer.
\newblock Game-theoretic statistics and safe anytime-valid inference.
\newblock \emph{Statistical Science}, 2023.

\bibitem[Rezatofighi et~al.(2019)Rezatofighi, Tsoi, Gwak, Sadeghian, Reid, and Savarese]{giou}
Hamid Rezatofighi, Nathan Tsoi, JunYoung Gwak, Amir Sadeghian, Ian Reid, and Silvio Savarese.
\newblock Generalized intersection over union: A metric and a loss for bounding box regression.
\newblock \emph{Conference on Computer Vision and Pattern Recognition}, 2019.

\bibitem[Rudenko et~al.(2020)Rudenko, Palmieri, Herman, Kitani, Gavrila, and Arras]{rudenko2020human}
Andrey Rudenko, Luigi Palmieri, Michael Herman, Kris~M Kitani, Dariu~M Gavrila, and Kai~O Arras.
\newblock Human motion trajectory prediction: A survey.
\newblock \emph{The International Journal of Robotics Research}, 2020.

\bibitem[Sheng et~al.(2020)Sheng, Zhang, Wu, Wang, Lyu, Ke, and Xiong]{sheng2020hypothesis}
Hao Sheng, Yang Zhang, Yubin Wu, Shuai Wang, Weifeng Lyu, Wei Ke, and Zhang Xiong.
\newblock Hypothesis testing based tracking with spatio-temporal joint interaction modeling.
\newblock \emph{IEEE Transactions on Circuits and Systems for Video Technology}, 2020.

\bibitem[Shin et~al.(2018)Shin, Mou, Mou, and Wang]{vehicle-navigation}
Bok-Suk Shin, Xiaozheng Mou, Wei Mou, and Han Wang.
\newblock Vision-based navigation of an unmanned surface vehicle with object detection and tracking abilities.
\newblock \emph{Machine Vision and Applications}, 2018.

\bibitem[Shin et~al.(2020)Shin, Kim, Kim, and Paik]{tracking-failure-KCF}
Jungsup Shin, Heegwang Kim, Dohun Kim, and Joonki Paik.
\newblock Fast and robust object tracking using tracking failure detection in kernelized correlation filter.
\newblock \emph{Applied Sciences}, 2020.

\bibitem[Shin et~al.(2023)Shin, Ramdas, and Rinaldo]{e-detectors}
Jaehyeok Shin, Aaditya Ramdas, and Alessandro Rinaldo.
\newblock E-detectors: a nonparametric framework for sequential change detection.
\newblock \emph{The New England Journal of Statistics in Data Science}, 2023.

\bibitem[Smith et~al.(2005)Smith, Gatica-Perez, Odobez, and Ba]{smith2005evaluating}
Kevin Smith, Daniel Gatica-Perez, J Odobez, and Sileye Ba.
\newblock Evaluating multi-object tracking.
\newblock \emph{Conference on Computer Vision and Pattern Recognition Workshops}, 2005.

\bibitem[Sun et~al.(2020)Sun, Cao, Jiang, Zhang, Xie, Yuan, Wang, and Luo]{sun2020transtrack}
Peize Sun, Jinkun Cao, Yi Jiang, Rufeng Zhang, Enze Xie, Zehuan Yuan, Changhu Wang, and Ping Luo.
\newblock Transtrack: Multiple object tracking with transformer.
\newblock \emph{arXiv preprint arXiv:2012.15460}, 2020.

\bibitem[Timans et~al.(2025)Timans, Verma, Nalisnick, and Naesseth]{timans2025continuous}
Alexander Timans, Rajeev Verma, Eric Nalisnick, and Christian~A. Naesseth.
\newblock On continuous monitoring of risk violations under unknown shift.
\newblock \emph{Conference on Uncertainty in Artificial Intelligence}, 2025.

\bibitem[Ville(1939)]{ville1939etude}
Jean Ville.
\newblock \emph{Etude critique de la notion de collectif}.
\newblock Gauthier-Villars Paris, 1939.

\bibitem[Vovk(2021)]{vovk2021testing}
Vladimir Vovk.
\newblock Testing randomness online.
\newblock \emph{Statistical Science}, 2021.

\bibitem[Vovk and Wang(2024)]{merging-seq-e-values}
Vladimir Vovk and Ruodu Wang.
\newblock Merging sequential e-values via martingales.
\newblock \emph{Electronic Journal of Statistics}, 2024.

\bibitem[Wald(1945)]{wald}
A. Wald.
\newblock Sequential tests of statistical hypotheses.
\newblock \emph{The Annals of Mathematical Statistics}, 1945.

\bibitem[Wang et~al.(2017)Wang, Liu, and Huang]{object-tracking-apce}
Mengmeng Wang, Yong Liu, and Zeyi Huang.
\newblock Large margin object tracking with circulant feature maps.
\newblock \emph{Conference on Computer Vision and Pattern Recognition}, 2017.

\bibitem[Wang and Ramdas(2022)]{wang2022false}
Ruodu Wang and Aaditya Ramdas.
\newblock False discovery rate control with e-values.
\newblock \emph{Journal of the Royal Statistical Society Series B: Statistical Methodology}, 2022.

\bibitem[Waudby-Smith and Ramdas(2024)]{aGRAPA}
Ian Waudby-Smith and Aaditya Ramdas.
\newblock Estimating means of bounded random variables by betting.
\newblock \emph{Journal of the Royal Statistical Society Series B: Statistical Methodology}, 2024.

\bibitem[Wu et~al.(2015)Wu, Lim, and Yang]{otb2015}
Yi Wu, Jongwoo Lim, and Ming-Hsuan Yang.
\newblock Object tracking benchmark.
\newblock \emph{IEEE Transactions on Pattern Analysis and Machine Intelligence}, 2015.

\bibitem[Xu et~al.(2020)Xu, Feng, Wu, and Kittler]{afat}
Tianyang Xu, Zhenhua Feng, Xiao{-}Jun Wu, and Josef Kittler.
\newblock {AFAT:} adaptive failure-aware tracker for robust visual object tracking.
\newblock \emph{arXiv preprint arXiv:2005.13708}, 2020.

\bibitem[Yan et~al.(2021)Yan, Peng, Fu, Wang, and Lu]{yan2021learning}
Bin Yan, Houwen Peng, Jianlong Fu, Dong Wang, and Huchuan Lu.
\newblock Learning spatio-temporal transformer for visual tracking.
\newblock \emph{Conference on Computer Vision and Pattern Recognition}, 2021.

\bibitem[Yao et~al.(2025)Yao, Guo, Yan, Ren, and Cao]{Yao_2025}
Siyuan Yao, Yang Guo, Yanyang Yan, Wenqi Ren, and Xiaochun Cao.
\newblock Unctrack: Reliable visual object tracking with uncertainty-aware prototype memory network.
\newblock \emph{IEEE Transactions on Image Processing}, 2025.

\bibitem[Ye et~al.(2022)Ye, Chang, Ma, Shan, and Chen]{ye2022joint}
Botao Ye, Hong Chang, Bingpeng Ma, Shiguang Shan, and Xilin Chen.
\newblock Joint feature learning and relation modeling for tracking: A one-stream framework.
\newblock \emph{European Conference on Computer Vision}, 2022.

\bibitem[Zhen et~al.(2020)Zhen, Fei, Wang, and Du]{zhen2020visual}
Xinxin Zhen, Shumin Fei, Yinmin Wang, and Wei Du.
\newblock A visual object tracking algorithm based on improved tld.
\newblock \emph{Algorithms}, 2020.

\end{thebibliography}
}

\clearpage
\onecolumn
\appendix
\begin{center}
    {\Large\bfseries Detecting Object Tracking Failure via Sequential Hypothesis Testing} \\[0.5em]
    {\Large\bfseries --- Supplementary Material ---}
\end{center}

\newtheorem{theorem}{Theorem}[section]
\newtheorem{corollary}{Corollary}[theorem]
\newtheorem{lemma}[theorem]{Lemma}
\newtheorem{definition}[theorem]{Definition}
\newtheorem{proposition}[theorem]{Proposition}
\newtheorem{example}[theorem]{Example}

\section{Background knowledge about Hypothesis Testing and E-processes}
\label{sec:ap-e-processes}

\subsection{Hypothesis Testing}
Hypothesis testing is the statistical method we adopt to provide formal guarantees in object tracking. This section introduces the foundational concepts of hypothesis testing and p-values, and then presents the more recent e-value framework as a first step to develop e-processes in the next section. The definitions presented in this section are adapted from~\cite{e-book, wang2022false}.

\subsubsection{Classical Hypothesis Testing and p-values}
Hypothesis testing~\cite{fisher1970statistical} is a fundamental statistical method used to make inferences about populational distributions based on sample data. It is a widely used technique in scientific research, quality control, and decision-making processes to draw conclusions based on sample evidence.

The process involves formulating two competing hypotheses:
\begin{itemize}
    \item The \textbf{null hypothesis} ($H_0$): a statement that there is no effect or no difference, and any observed variation is due to random chance.
    \item The \textbf{alternative hypothesis} ($H_1$): a statement that contradicts the null hypothesis, suggesting that there is a true effect or difference.
\end{itemize}

The goal of hypothesis testing is to assess whether the observed data provides sufficient evidence to reject the null hypothesis in favor of the alternative. This is typically done by calculating a \textbf{test statistic} (a function of the observed data) and comparing it to a threshold determined by a pre-specified \textbf{significance level} ($\alpha$), often set at 0.01, 0.05 or 0.1.

P-values are a widely used measure of statistical evidence in hypothesis testing. They are defined as the probability of observing a statistic as extreme as the one obtained, assuming the null hypothesis $H_0$ is true. Therefore, \textbf{a small p-value indicates strong evidence against $H_0$}. In practice, the decision of whether $H_0$ should be rejected is made by comparing the p-value to a predefined significance level (commonly $\alpha = 0.01$, $0.05$, or $0.1$). If the p-value is lower than $\alpha$, the result is considered statistically significant. Conversely, a large p-value suggests that the observed data is consistent with $H_0$, and we fail to reject it (note that rejecting $H_0$ does not imply $H_1$ is true — it simply means we lack sufficient evidence against $H_0$).

Example \ref{ex:p-test-gaussian} outlines a basic hypothesis test on the mean of a normal distribution. 

 \begin{example}
    \label{ex:p-test-gaussian}
    We observe $n$ i.i.d. samples $X_1, X_2, \dots, X_n$ from a normal distribution with unknown mean $\mu$ and known variance $\sigma^2$. We want to test whether the mean is equal to a specified value $\mu_0$:
    \begin{equation*}
        H_0: \mu = \mu_0 ; \quad H_1: \mu \neq \mu_0.
    \end{equation*}
    Under $H_0$, the sample mean is normally distributed:
    \begin{equation*}
    \bar{X}_n \sim \mathcal{N}(\mu_0, \sigma^2/n).
    \end{equation*}
    We define the standardized test statistic:
    \begin{equation*}
    Z = \frac{\bar{X}_n - \mu_0}{\sigma / \sqrt{n}} \sim \mathcal{N}(0,1),
    \end{equation*}
    Given a significance level $\alpha \in (0,1)$ (commonly $\alpha = 0.05$), we reject the null hypothesis if the test statistic falls in the critical region:
    \begin{equation*}
    \mathcal{R} = \{|Z| > z_{1 - \alpha/2} \},
    \end{equation*}
    where $z_{1 - \alpha/2}$ is the $(1 - \alpha/2)$ quantile of the standard normal distribution.
    \begin{center}
    \includegraphics[width=0.3\linewidth]{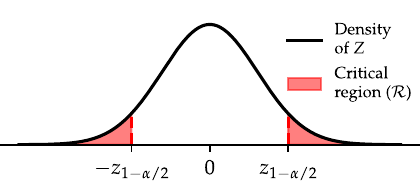}
    \end{center}
    In other words, \textbf{we reject the null if the statistic falls in the most extreme $\boldsymbol \alpha$\% fraction of its distribution under the null}.\\
    Alternatively, we compute the $p$-value:
    \begin{equation*}
    p = 2 \cdot \mathbb{P}(Z > |z|),
    \end{equation*}
    where $z$ is the observed value of the statistic, and reject $H_0$ if $p < \alpha$.
\end{example}

Let's now formalize the hypothesis testing framework. We begin by defining hypotheses as sets of probability measures:

\begin{definition}
    Given a measurable space $(\Omega, \mathcal{F})$, a \textbf{hypothesis} is a set of probability measures on $(\Omega, \mathcal{F})$.
    
    Usually, we use the following notation for a hypothesis test:
    \begin{equation}
        H_0: \mathbb{P} \in \mathcal P; \quad H_1: \mathbb{Q} \in \mathcal Q,
    \end{equation}
    with $\mathcal P \cap \mathcal Q = \emptyset$.
    
    We say that a hypothesis is \textbf{simple} if $|\mathcal{P}| = 1$ and \textbf{composite} otherwise.
\end{definition}

To evaluate evidence in data, we define tests:

\begin{definition}
    A binary test $\phi$ is a $\{0,1\}-$valued random variable.
\end{definition}

If the test takes the value $\phi = 1$, we reject the null hypothesis, while $\phi = 0$ means not rejecting it. We can compute the \textbf{Type-I error} or \textbf{false positive rate} of a test $\phi$ for a distribution $\mbp \in \mathcal{P}$ as $\expec{\mbp}{\phi}$. That is, the probability of rejecting the null hypothesis when it is actually true. For a composite hypothesis $\mcp$, we say that the test has level $\alpha$ if its Type-I error is at most $\alpha$ for every $\mbp \in \mcp$.

Let's now formalize the concept of p-values.

\begin{definition}
    A \textbf{p-variable} $P$ for $\mcp$ is a $[0, \infty)-$ valued random variable such that $\prob{P \leq \alpha} \leq \alpha$ for all $\alpha \in (0, 1)$ and all $\mbp \in \mcp$.
    \begin{itemize}
        \item If the equality holds in the condition above ($\prob{P \leq \alpha} = \alpha$), we say that $P$ is an \textbf{exact} p-variable.
    \end{itemize}
\end{definition}

We use the term \textbf{p-value} to refer to the realized value of a p-variable computed from observed data\footnote{In practice, "p-value" is used for both concepts, even though this is an abuse of notation.}.  Intuitively, for usual significance levels $\alpha$, small p-values are unlikely under the null hypothesis. Therefore, \textbf{a low p-value gives statistical evidence against $H_0$}, while a high p-value indicates that the data is consistent with $H_0$, and there is insufficient evidence to reject it.

In practice, p-variables are usually computed as the cumulative distribution function of a test statistic, as shown in.

\begin{example}
\label{ex:p-value}
Let's go back to Example \ref{ex:p-test-gaussian} to better understand how the p-variable is constructed. We defined the p-variable $P$ as the probability that a standard normal variable exceeds the observed test statistic in magnitude:
$$P = 2 \cdot \mathbb{P}(Z > |Z_\text{obs}|), \quad  Z \sim \mathcal{N}(0, 1).$$
\begin{center}
    \includegraphics[width=0.3\linewidth]{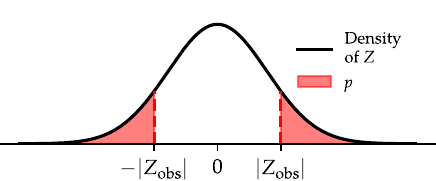}
\end{center}

Note that $P$ is a function of the test statistic $Z$.
Since $ Z_{\text{obs}} $ under $ H_0 $ also follows $ \mathcal{N}(0,1) $, we have:
\begin{equation*}
\begin{array}{l}
\mathbb{P}_{H_0}(P \leq \alpha) \\
\quad = \mathbb{P}_{H_0}\left(2 \cdot \mathbb{P}(Z > |Z_{\text{obs}}|) \leq \alpha \right) \\
\quad = \mathbb{P}_{H_0}\left(|Z_{\text{obs}}| \geq z_{1 - \alpha/2} \right) \\
\quad = \alpha.
\end{array}
\end{equation*}
Hence, $ P $ is an \emph{exact} p-variable.
\end{example}

\subsubsection{Hypothesis Testing with e-values}
Opposed to the classical p-value framework, hypothesis testing can also be performed using \textbf{e-values}~\cite{e-book}, which offer a different and often more flexible approach to measuring evidence against a null hypothesis. E-values are built on expected values rather than tail probabilities. While p-values quantify how extreme the data are under the null, e-values quantify how much evidence the data provide for the alternative hypothesis, relative to the null.\\

We can formally define an e-variable as follows:

\begin{definition}
    An \textbf{e-variable} $E$ for $\mcp$ is a $[0, \infty]-$ valued random variable such that $\expec{\mbp}{E} \leq 1$ for all $\mbp \in \mcp$.
    \label{def:e-value}
\end{definition}
Contrary to p-variables, e-variables are expected to be small under the null. Therefore, high e-values indicate statistical evidence against $H_0$.

Example \ref{ex:e-test-gaussian} shows a concrete construction of an e-variable in the Gaussian setting, using a likelihood ratio between simple hypotheses.

\begin{example}
    \label{ex:e-test-gaussian}
    We consider the same data as in Example \ref{ex:p-test-gaussian}, where we observe $n$ i.i.d. samples from a Gaussian distribution with known variance $\sigma^2$, but we now aim to test:
    \begin{equation*}
        H_0: \mu = \mu_0, \quad H_1: \mu = \mu_1,
    \end{equation*}
    for some fixed alternative $\mu_1 \neq \mu_0$.
    Instead of a p-variable, we construct an \textbf{e-variable} using a likelihood ratio between the alternative $\mu_1$ and the null $\mu_0$:
    {\begin{align*}
    E &= \frac{f(X_1, \dots, X_n; \mu_1)}{f(X_1, \dots, X_n; \mu_0)} \\
      &= \prod_{i=1}^n \frac{f(X_i; \mu_1)}{f(X_i; \mu_0)} \\
      &= \exp\left(
          \frac{1}{2\sigma^2} \sum_{i=1}^n \left[
              (X_i - \mu_0)^2 - (X_i - \mu_1)^2
          \right]
        \right) \\
      &= \exp\left(
          \frac{n}{2\sigma^2} \left[
              2(\mu_1 - \mu_0)\bar{X}_n + (\mu_0^2 - \mu_1^2)
          \right]
        \right)
    \end{align*}}
    This ratio compares the likelihood under a fixed alternative $\mu_1$ to that under the null $\mu_0$. It satisfies the condition to be an (exact) e-variable:
    \begin{align*}
    \mathbb{E}_{H_0}[E]
    &= \mathbb{E}_{\mu_0}\left[ \frac{f(\mathbf{x}; \mu_1)}{f(\mathbf{x}; \mu_0)} \right]\\
    &= \int \frac{f(\mathbf{x}; \mu_1)}{f(\mathbf{x}; \mu_0)} f(\mathbf{x}; \mu_0)\, d\mathbf{x} \\
    &= \int f(\mathbf{x}; \mu_1)\, d\mathbf{x} = 1.
    \end{align*}
    We reject $H_0$ if the e-variable exceeds a threshold $1/\alpha$:
    \begin{equation*}
    E > \frac{1}{\alpha},
    \end{equation*}
    which yields a test that controls the Type-I error at level $\alpha$.
\end{example}

While classical hypothesis testing provides a rigorous framework for decision-making under uncertainty, it typically assumes a fixed dataset and does not account for scenarios where data arrives over time. However, many real-world applications such as object tracking require decisions to be made sequentially as new observations are collected. To address this, we now turn to sequential hypothesis testing, which extends classical methods to dynamic, time-evolving settings while maintaining statistical validity at each step.

\subsection{Sequential Hypothesis Testing}
\label{app:seq-testing}

In classical hypothesis testing, the number of samples is fixed in advance: we collect a dataset of size $n$, compute a test statistic, and make a binary decision to reject or retain the null hypothesis. However, in many real-world settings such as streaming data, online learning, or adaptive systems like object tracking this assumption does not hold. In sequential hypothesis testing, we receive data one point at a time and want to decide \emph{on the fly} whether we have enough evidence to reject the null or whether we should keep receiving data.

Sequential testing introduces a fundamental challenge: if we monitor a test statistic such as a p-value or e-value repeatedly over time and stop the test when it appears significant, we risk inflating the probability of false discoveries. This issue arises because classical p-values and fixed-time e-values are not generally valid when the stopping time is data-dependent. 

To address this, we must use tools designed for sequential settings. In particular, e-processes (sequences of e-values structured as nonnegative supermartingales) offer a robust solution. They remain valid under arbitrary stopping rules and allow continuous monitoring while preserving rigorous Type-I error guarantees. In this section, we introduce the basic mathematical machinery behind these tools. The definitions presented in this section are adapted from~\cite{e-book}.

\subsubsection{Random Processes and Martingales}

To define sequential tests, we work with sequences of random variables that evolve over time. These are called \textbf{random processes} and are defined as follows:

\begin{definition}
    We consider a sample space $\Omega$ and a filtration $\mathbf{F} = (\mathcal{F}_t)_{t \geq 0}$, representing the information available up to each time $t$. A sequence of random variables $(W_t)_{t \geq 0}$ is a \textbf{process} adapted to $\mathbf{F}$ if $W_t$ is measurable with respect to $\mathcal{F}_t$ for all $t$. Usually, we take $\mathcal{F}_t$ to be the natural filtration of the observed data: $\mathcal{F}_t = \sigma(W_1, ..., W_t)$.
\end{definition}

There is a special type of random process that is relevant for our purpose: \textbf{martingales}. These can be thought of as fair game processes, where the expected future value is equal to the current one, given the past.

\begin{definition}
    A process $\{W_t\}_{t \geq 0}$ is a \textbf{martingale} with respect to a filtration $(\mathcal{F}_t)$ if:
    \begin{enumerate}[(i)]
        \item $\mathbb{E}[W_t] < \infty$ for all $t$;
        \item $\mathbb{E}[W_t \mid \mathcal{F}_{t-1}] = X_{t-1}$ for all $t \geq 1$.
    \end{enumerate}
    If condition (2.) holds with ``$\leq$'' instead of ``$=$'', we say that $W_t$ is a \textbf{supermartingale}.
\end{definition}

Supermartingales play a central role in safe anytime-valid inference (SAVI) because they enable the construction of \textbf{e-processes}: sequences of e-values that remain valid under arbitrary stopping.

In sequential analysis, we may want to stop the test early if we accumulate strong evidence, or continue testing if evidence is weak. To formalize this, we introduce the notion of a \textbf{stopping time}:

\begin{definition}
    A \textbf{stopping time} (or \textbf{stopping rule}) $\tau$ is a nonnegative, integer-valued random variable such that the event $\{\tau \leq t\}$ is measurable with respect to $\mathcal{F}_t$ for all $t \geq 0$. That is, at time $t$, we can make a decision whether we should stop or continue.
\end{definition}

From now on, we denote the set of all stopping times as $\mathcal{T}$.

In traditional sequential tests (like Wald's Sequential Probability Ratio Test~\cite{wald}), stopping rules must be predefined. However, e-processes remain valid even if we pick $\tau$ adaptively based on the data.

\subsubsection{Sequential Tests}
Let's now formalize the notion of sequential test
\begin{definition}
    A level-$\alpha$ \textbf{sequential test} for $\mcp$ is a binary process $\phi$ such that
    \begin{equation}
    \label{eq:seq_test_def}
        \prob{\exists t \geq 1 : \phi_t = 1} \leq \alpha \text{ for all } \mbp \in \mcp.
    \end{equation}

\end{definition}

We can also identify a test by the stopping time $\tau := \inf \{t \geq 1 : \phi_t = 1\}$ (with the convention $\inf \emptyset = \infty$). Then, the condition in Equation \autoref{eq:seq_test_def} turns into 
\begin{equation}
    \expec{\mbp}{\phi_\tau} \leq 1 \text{ for all } \forall \mbp \in \mcp, \tau \in \mathcal T
\end{equation}

In other words, a sequential test is a decision rule that can monitor data sequentially and raise an alarm (i.e., reject the null) at any point in time, but still guarantees that the probability of making a false discovery remains below $\alpha$, no matter when we choose to stop. This is a much stronger requirement than in classical hypothesis testing, because it ensures validity even if the stopping time is random, data-dependent, or not specified in advance. 

\subsubsection{E-processes}
E-processes provide such guarantees by generalizing the concept of e-values to entire processes:

\begin{definition}
    A non-negative process $\{W_t \}_{t \in \mathcal{T}}$ that is adapted to $\mathcal{F}$ is called an \textbf{e-process} if 
    \begin{equation}
        \expec{\mbp}{W_\tau} \leq 1 \text{ for any stopping time } \tau \in \mathcal{T} \text{ and any } \mbp \in \mcp.
    \end{equation}
\end{definition}
This definition ensures that the process controls Type-I error uniformly over time. In particular, if the null hypothesis holds, then the e-process cannot grow too large on average, regardless of when we choose to stop and make a decision.


Now, we will introduce a relevant result about e-processes that allows to build sequential tests from them with statistical guarantees on the Type-I error~\cite{ville1939etude}:
\begin{proposition}
    \textbf{(Ville's inequality)} If $\{W_t\}_{t \in \mathcal{T}}$ is an E-process for a null $\mcp$, then for any $\alpha \geq 1$:
    \begin{equation}
        \sup_{P \in \mcp} P \left( \sup_t W_t \geq \frac{1}{\alpha} \right) \leq \alpha \; \; \forall \alpha \in (0, 1).
    \end{equation}
    \label{prop:ville}
\end{proposition}

Although we are not including the proof of this result, note that it is just a generalization to e-processes of Markov's inequality applied to e-variables. Given an e-variable $E$, applying Markov's inequality and the definition of e-value gives:
\begin{equation*}
    \prob{E \geq \alpha} \leq \frac{\expec{}{E}}{\alpha} \leq \frac{1}{\alpha}.
\end{equation*}
This result allows us to obtain a sequential test from an e-process by rejecting the null as soon as the process exceeds $\frac{1}{\alpha}$ \textit{at any time}, with the guarantee that Type-I error is upper bounded by $\alpha$:

\begin{equation}
    \tau = \inf\left\{t: W_t \geq \frac{1}{\alpha}\right\}
\end{equation}

To apply sequential hypothesis testing in practice, we require a principled method for constructing e-processes—test statistics that grow when the data is incompatible with the null and remain controlled otherwise. While the definition of an e-process ensures validity under arbitrary stopping, it does not prescribe how to design such a process. Indeed, this is the key challenge driving much of the recent work: how to construct suitable e-processes for increasingly complex composite hypotheses, particularly in the context of deep learning and adaptive models. In the following section, we present an approach for building e-processes by modeling statistical inference as a dynamic betting game.

\subsubsection{Constructing e-processes through Testing by Betting}
\label{subsec:testing-by-betting}
E-processes can be interpreted as a betting game between a statistician and nature~\cite{aGRAPA, ramdas2023game, e-book}. The statistician starts with an initial wealth and sequentially places bets $B_t: \mathcal{W} \rightarrow [0, \infty]$ based on past observations, such that the expected value under the null is at most one:
\begin{equation}
    \label{eq:bet}
    \expec{\mbp}{B_t | W_1, ..., W_{t-1}} \leq 1 \; \; \forall \mbp \in \mcp.
\end{equation}

The wealth process is initialized with $W_0 = 1$ and sequentially built as 
\begin{equation}
    \label{eq:wealth}
    W_t = W_{t-1} \cdot B_t.
\end{equation}

If the null hypothesis is true, the condition in \autoref{eq:bet} implies that no betting strategy should consistently increase wealth defined in \autoref{eq:wealth}. However, under an alternative $\mbq \in \mcq$, it is possible to increase wealth systematically. Therefore, \textbf{the wealth process $\{W_t\}_{t \in \mathcal{T}}$ can be interpreted as a measure of the evidence against the null}.

Following the testing-by-betting framework, we can construct an \textbf{empirically adaptive e-process} by combining several e-variables $\{E_t\}_{t \in \mathcal{T}}$ via the following martingale:
\begin{equation}
    W_0 = 1, \; \; W_t = \prod_{i=1}^t ((1 - \lambda_i) + \lambda_i E_i).
    \label{eq:empirically-adaptive-e-process}
\end{equation}

Here the bet at each timestep $t$ is $B_t = (1 - \lambda_t) + \lambda_t E_t$, where $\lambda_t \in [0, 1]$ (the betting rate) is a measurable function of $E_1, ..., E_{t-1}$.

To build this process, we need to choose a betting rate $\lambda_t$ at each timestep. We do this by aiming for the optimal e-process that grows as fast as possible under the alternative \footnote{This is called log-optimality in the e-values literature \cite{e-book}.} We can achieve this by maximizing the wealth (the bet evaluated at the realized value of $W_t$) at each timestep:
\begin{equation}
    \lambda_t = \arg\max_{\lambda \in [0, 1]} \expec{H_1}{\log W_t \mid \mathcal{F}_{t-1}}
\end{equation}

Solving this optimization problem implies a significant computational expense, and therefore we usually opt for numerical approximations such as GRAPA \cite{aGRAPA} or SF-OGD \cite{sf-ogd}, presented in \autoref{subsec:method-bettingrate}.

\section{Derivation of the Empirically Adaptive e-process for Object Tracking Failure Detection}
\label{sec:ap-derivation}
Our goal is to solve the hypothesis test in \autoref{eq:hypothesis} by building an e-process $\{X_t\}_{t \ge 0}$. 

We leverage the empirically adapted e-process framework presented in \autoref{subsec:testing-by-betting} to build an e-process for this test. For each timestep, we can define an e-value based on the discrepancy between the tracking metric at frame $t$ and the threshold:

\begin{equation}
    E_t = 1 + \varepsilon - m_t.
    \label{eq:e-value-tracking-failure}
\end{equation}

By the linearity of the expectation and Equation \autoref{eq:hypothesis}, $E_t$ meets Definition \ref{def:e-value} and therefore is an e-value. 

We can use the empirically adaptive e-process defined in \autoref{eq:empirically-adaptive-e-process} to merge these e-values:

\begin{align}
    W_t &= \prod_{i=1}^t ((1 - \lambda_i) + \lambda_i E_i) \\
    &= \prod_{i=1}^t ((1 - \lambda_i) + \lambda_i (1 + \varepsilon - m_i))\\\
    &= \prod_{i=1}^t (1 + \lambda_i (\varepsilon - m_i)),
\end{align}

which results in the process defined in \autoref{eq:e-process}.

\end{document}